\documentclass[10pt,twocolumn,letterpaper]{article}

\usepackage{amsmath,amsfonts}
\usepackage{algorithmic}
\usepackage{algorithm}
\usepackage{array}
\usepackage[caption=false,font=normalsize,labelfont=sf,textfont=sf]{subfig}
\usepackage{textcomp}
\usepackage{stfloats}
\usepackage{verbatim}
\usepackage{graphicx}
\usepackage{amssymb}
\usepackage{multirow}
\usepackage[table,xcdraw]{xcolor}
\usepackage{color}
\usepackage{amssymb}
\usepackage{pifont}
\usepackage{booktabs}
\usepackage[normalem]{ulem}
\useunder{\uline}{\ul}{}

\usepackage[pagenumbers]{iccv} 

\definecolor{iccvblue}{rgb}{0.21,0.49,0.74}
\usepackage[pagebackref,breaklinks,colorlinks,allcolors=iccvblue]{hyperref}

\def\paperID{*} 
\def\confName{ICCV}
\def\confYear{2025}

\title{Beyond Single Images: Retrieval Self-Augmented Unsupervised \\ Camouflaged Object Detection}

\author{Ji Du$^{1,2}$, Xin Wang$^{2}$, Fangwei Hao$^{1,*}$, Mingyang Yu$^{1}$ \\
Chunyuan Chen$^{1}$, Jiesheng Wu$^{3}$, Bin Wang$^{1}$, Jing Xu$^{1,*}$, Ping Li$^{2,}$\thanks{Corresponding authors: Fangwei Hao, Jing Xu (xujing@nankai.edu.cn) and Ping Li (p.li@polyu.edu.hk).}\\
$^1$College of Artificial Intelligence, Nankai University, China \\
$^2$Department of Computing, The Hong Kong Polytechnic University, Hong Kong \\
$^3$School of Computer and Information, Anhui Normal University, China
}

\begin{document}
\maketitle

\begin{abstract}
At the core of Camouflaged Object Detection (COD) lies segmenting objects from their highly similar surroundings. Previous efforts navigate this challenge primarily through image-level modeling or annotation-based optimization. Despite advancing considerably, this commonplace practice hardly taps valuable dataset-level contextual information or relies on laborious annotations. In this paper, we propose RISE, a \textbf{R}etr\textbf{I}eval \textbf{SE}lf-augmented paradigm that exploits the entire training dataset to generate pseudo-labels for single images, which could be used to train COD models. RISE begins by constructing prototype libraries for environments and camouflaged objects using training images (without ground truth), followed by K-Nearest Neighbor (KNN) retrieval to generate pseudo-masks for each image based on these libraries. It is important to recognize that using only training images without annotations exerts a pronounced challenge in crafting high-quality prototype libraries. In this light, we introduce a Clustering-then-Retrieval (CR) strategy, where coarse masks are first generated through clustering, facilitating subsequent histogram-based image filtering and cross-category retrieval to produce high-confidence prototypes. In the KNN retrieval stage, to alleviate the effect of artifacts in feature maps, we propose Multi-View KNN Retrieval (MVKR), which integrates retrieval results from diverse views to produce more robust and precise pseudo-masks. Extensive experiments demonstrate that RISE outperforms state-of-the-art unsupervised and prompt-based methods. Code is available at \url{https://github.com/xiaohainku/RISE}.
\end{abstract}

\section{Introduction}

Camouflaged Object Detection (COD)~\cite{ref1} is dedicated to segmenting objects meticulously concealed within their surroundings.
Existing research efforts have predominantly concentrated on leveraging contextual information \textbf{within individual images} to accurately delineate and extract camouflaged objects from visually homogeneous backgrounds.

\begin{figure}[!t]
\centering
\begin{center}
\includegraphics[width=\linewidth]{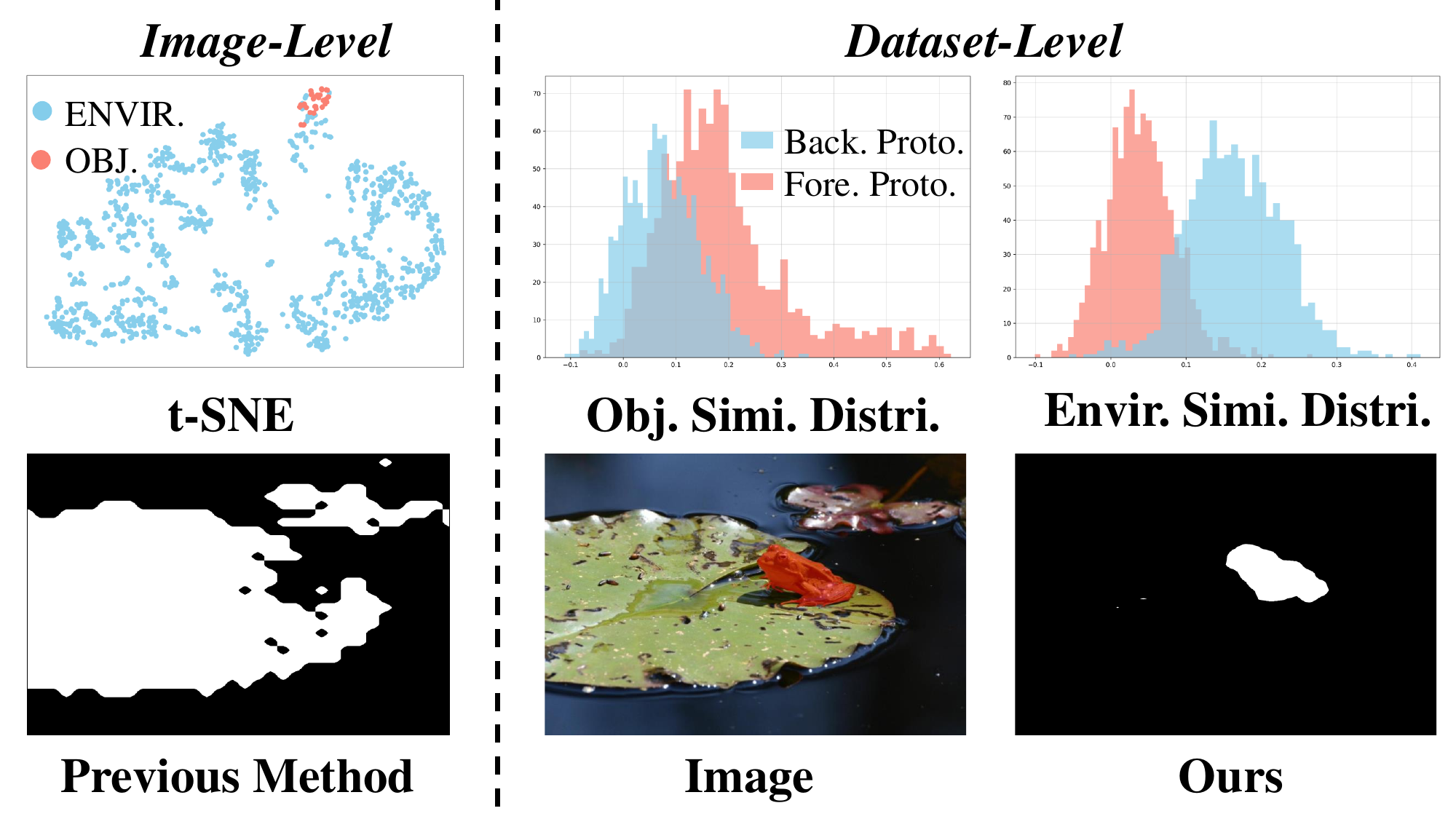}
\end{center}
\vspace{-1em}
\caption{\textbf{LEFT}: t-SNE visualization of DINOv2~\cite{oquab2023dinov2} features and segmentation results from previous unsupervised methods. The highly similar characteristics between the camouflaged object and its environment cause the intra-image similarity-based approach to underperform on the COD task. \textbf{RIGHT}: Similarity distributions of global features of camouflaged objects and environments on the respective foreground and background prototype libraries. These prototype libraries are constructed from the COD dataset. Both the camouflaged object and the environment show a higher similarity to their corresponding prototype libraries, suggesting that dataset-level information can be effectively utilized to distinguish between the highly similar foreground and background in a single image.}
\vspace{-1em}
\label{motivation}
\end{figure}

Mainstream supervised learning methods, including fully supervised~\cite{ref2, ugtr, zoomnet, segmar, ref5, vscode}, weakly supervised~\cite{wscod, wssam, wscod_point, sam-cod, Xu_Siu_Lau_2024}, and semi-supervised~\cite{PNet, CamoTeacher}, exploit varying degrees of \textbf{single-image-level} supervised signals, such as dense annotations, scribbles, bounding boxes, and categories, to model the intrinsic relationship between camouflaged objects and their environments. Despite advancements, these methods remain constrained, to varying degrees, by the time-consuming annotation process.

To address this limitation, prompt-based segmentation~\cite{gensam, mmcpf, ProMaC} has emerged as a promising alternative. This approach leverages pre-trained foundation models and task-specific prompt words to localize camouflaged objects, subsequently generating prompts for the Segment Anything Model (SAM)~\cite{sam} to perform segmentation. However, this paradigm still relies on some form of supervision and is fundamentally constrained by the limited camouflage-specific contextual understanding embedded \textbf{in a single image}.

Unsupervised COD, which further omits task-specific prompts compared to prompt-based segmentation, thus far represents a largely uncharted research territory. A straightforward practice is to migrate general unsupervised methods~\cite{tokencut, FOUND, deepspectral} to COD. However, as highlighted in~\cref{motivation}, these methods, which rely solely on feature similarity \textbf{within a single image}, often struggle to differentiate between the foreground and background due to the high similarity between the camouflaged object and its surroundings.

Both prompt-based segmentation and unsupervised approaches demonstrate substantial performance gaps compared to supervised methods, primarily attributable to insufficient COD-specific contextual understanding. This limitation manifests in two critical aspects: firstly, foundation models trained on general datasets are not optimized for the unique challenges posed by COD; more significantly, both paradigms focus predominantly on \textbf{intra-image relationships}, ignoring potentially valuable \textbf{dataset-level contextual information} that could significantly enhance the differentiation between foreground and background objects.

Without fine-tuning foundation models to enrich camouflage-specific contextual understanding, we would like to ask: \textit{how could dataset-level contextual information be effectively leveraged to segment targets in single images?}

In this light, we introduce RISE, a retrieval self-augmented paradigm designed to effectively utilize dataset-level contextual information. RISE segments camouflaged objects by retrieving prototypes from libraries derived directly from the COD dataset. Unlike conventional retrieval-augmented methods~\cite{karazija24diffusion,Barsellotti_2024_CVPR,Wang_2024_CVPR}, which rely on external data sources for prototype libraries, RISE extracts high-quality prototypes directly from the camouflaged dataset itself, in an annotation-free manner.

In the absence of annotations, we introduce Clustering-then-Retrieval (CR), a method to extract high-quality, noise-resistant prototypes for both the environment and camouflaged objects from available images. CR begins by generating a coarse mask for each image using spectral clustering. Based on this mask, CR then constructs prototypes for the camouflaged object and environment by retrieving features that exhibit the least similarity to the respective background or foreground regions.

During the retrieval stage, we use K-Nearest Neighbor (KNN) to identify feature categories. However, artifacts in the foundation model's feature maps may introduce noise into the retrieval results. While fine-tuning the model with new tokens is a common solution~\cite{dino_artifact}, we propose a simpler yet effective alternative: Multi-View KNN Retrieval (MVKR). Recognizing that the location of these artifacts can vary with the image's viewpoint, MVKR mitigates their impact by combining retrieval results from images captured from multiple viewpoints, without any fine-tuning.

Comprehensive experiments across a range of benchmark datasets substantiate the effectiveness of our proposed method in segmenting camouflaged objects from their highly similar surroundings. Additionally, our approach significantly reduces the time required to generate high-quality pseudo-masks. In contrast to prompt-based segmentation methods, which may take \textbf{days} to generate masks for the entire dataset (4,040 images), our method completes the task in \textbf{hours}, with far less GPU memory usage. We succinctly summarize the key contributions of our work as follows:
\begin{itemize}
    \item We present a new paradigm for unsupervised COD by leveraging the COD dataset to construct prototype libraries, which serve as the foundation for retrieving camouflaged objects. In contrast to existing methods that rely on the features of individual images, our approach harnesses the global information at the dataset level, enabling a more accurate capture of the subtle distinctions between camouflaged objects and their background.
    \item We propose CR to extract environmental and camouflaged object prototypes from COD datasets. CR utilizes available camouflaged images, rather than external ones, and incorporates a unique joint clustering and retrieval mechanism to distinguish and mine prototypes, eliminating the need for manual annotations. Based on the prototype libraries, we propose Multi-View KNN Retrieval to segment camouflaged objects from the environment.
    \item Our extensive experiments validate the effectiveness of the proposed method.
\end{itemize}

\section{Related Work}
\subsection{Camouflaged Object Detection}
Camouflaged Object Detection has garnered growing research interest credited to its unique challenges and vast utility~\cite{fan2023advances, xiao2024survey}.
Existing methods can be categorized based on their degree of dependence on annotations into fully supervised learning, weakly supervised learning, prompt-based segmentation, and unsupervised learning.

As the most fully developed and mature branch, advances in fully supervised methods have consistently embraced the theme of maximizing the use of dense annotations. By developing camouflage-specific modules~\cite{ref1, distractionmining, mgl, ref2, preynet, mei2023camouflaged, hitnet, feder, fsp, camoformer, 10684046, ref7, 10537109, hao5089840distribution}, incorporating additional supervised signals (boundary~\cite{bsanet, sun2022bgnet, 10918801}, frequency~\cite{ref6, FaCOD,finet}, depth~\cite{ref5}, category~\cite{ovcod, zhao2025open}), or adopting innovative learning paradigms (joint modeling~\cite{lv2021simultaneously, jscod}, uncertainty-guided learning~\cite{ugtr}, bio-inspired mechanism~\cite{segmar, zoomnet}, unfolding network~\cite{he2025run}, generation-based strategy~\cite{ref8, camodiffusion1, iceg+, zhao2024camouflaged, 10834569, zhang2023camouflaged}, general model~\cite{evp,zhao2024towards, vscode, Spider}), fully supervised learning seeks to effectively model the relationship between camouflaged objects and their environment based on manual labeling. However, this advancement comes with a non-negligible labeling cost, as labeling a single image can take up to an hour~\cite{ref1}.

To alleviate the burden of pixel-level dense annotations, weakly supervised methods leverage easily accessible sparse annotations—such as scribbles~\cite{wscod}, bounding boxes~\cite{PNet}, points~\cite{wscod_point}, or categories~\cite{wscod_category}—to segment camouflaged objects in a cost-effective manner. This is achieved through carefully designed loss functions~\cite{wscod}, refined pseudo-masks from pre-trained segmentation models~\cite{wssam, see-wscod}, or bio-inspired detection processes~\cite{wscod_point}. To further reduce the need for labeling, another line of research, prompt-based segmentation~\cite{gensam, mmcpf, ProMaC}, explores the ``foundation models + SAM" paradigm for training-free COD using task-level prompt words. However, both types of methods exhibit significant performance gaps compared to fully supervised methods, due to limited information in sparse annotations or limited generalization abilities of foundation models when applied to camouflage contexts rather than general scenarios~\cite{Liu_2025_CVPR}.

Compared to previous methods, unsupervised approaches are particularly challenging, as they do not rely on labels or task-specific prompt words, making them an underexplored area in COD. Conventional unsupervised methods~\cite{LOST, deepspectral, FOUND, Tian_2024_CVPR} typically differentiate between foreground and background based on feature similarity within a single image. A common paradigm~\cite{tokencut, cutler, CuVLER, diffcut} is to construct a graph where features serve as nodes and the similarities between them form the edges, then apply Normalized Cut~\cite{ncut} to the graph for node segmentation. However, the highly similar characteristics of foreground and background in a single image cause these unsupervised methods, which perform well on general images, to underperform in COD.

Our approach pertains to unsupervised COD. Unlike traditional unsupervised methods that rely on image-level feature similarity, our proposed RISE focuses on dataset-level similarity and seeks to utilize cross-image information to separate camouflaged objects from the background.

\subsection{Retrieval-Augmented Methods}
In the context of natural language processing, Retrieval-Augmented Generation (RAG)~\cite{pmlr-v162-borgeaud22a} aims to enhance the accuracy and richness of generated content by combining information retrieval and generative models to retrieve relevant information from external knowledge bases.
In the field of computer vision, the emergence of foundation models, especially self-supervised~\cite{DINO, oquab2023dinov2} and generative models~\cite{ddpm, stablediffusion, he2025diffusion}, has spawned a new paradigm for segmentation: retrieval-augmented semantic segmentation~\cite{karazija24diffusion,Barsellotti_2024_CVPR,Wang_2024_CVPR}. This line of research focuses on crafting prototype libraries of different categories through these foundation models and accomplishing segmentation of specific categories through retrieving from the prototype libraries.

Our proposed retrieval self-augmented method, RISE, draws parallels to retrieval-augmented semantic segmentation. However, instead of relying on external models to generate prototype libraries, our approach extracts high-quality prototypes from the COD dataset itself through our unique clustering-then-retrieval mechanism.

\section{Method}
\begin{figure*}[!t]
\centering
\begin{center}
\includegraphics[width=\linewidth]{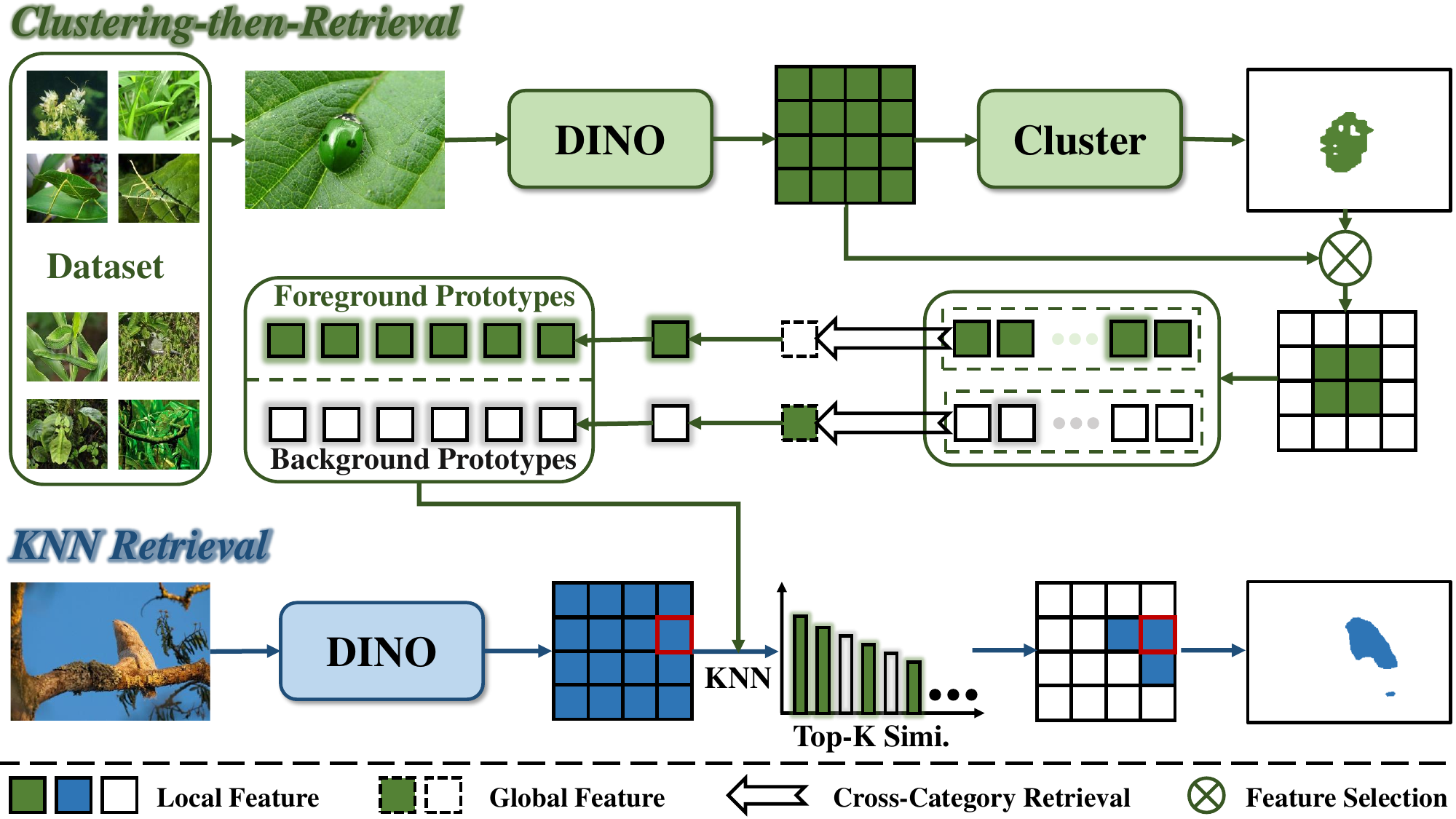}
\end{center}
\vspace{-1em}
\caption{Overview of RISE. RISE is composed of two main stages: Clustering-then-Retrieval (CR) and KNN Retrieval. These stages work together to generate prototype libraries and retrieve camouflaged objects. In the CR phase, we start by clustering the feature maps from DINO for each image in the dataset, which generates a coarse mask. This mask allows us to extract both local and global features for the foreground and background. Using cross-category retrieval, we then retrieve high-quality prototypes from the local features. By aggregating the prototypes across all images, we create the final foreground and background prototype libraries. In the second stage, KNN retrieval, we apply KNN to identify the top-$K$ most similar prototypes in prototype libraries for each feature in the feature map. A voting mechanism is then used to classify each feature as either foreground or background.
}
\vspace{-1em}
\label{overview}
\end{figure*}

In this paper, we tackle the problem of unsupervised camouflaged object detection with a simple retrieval self-augmented pipeline. Our work draws inspiration from conventional unsupervised segmentation methods~\cite{FOUND, tokencut}, which capitalize on the similarity between self-supervised representations to segment objects. While these methods primarily focus on feature similarity within a single image, our proposed retrieval-based approach leverages dataset-level information to effectively segment camouflaged objects. The overview of our retrieval self-augmented pipeline is illustrated in~\cref{overview}. First, we introduce Clustering-then-Retrieval (CR), a method that generates prototype libraries for camouflaged objects and environments using the COD dataset (\cref{sec:cr}). Second, leveraging the prototype libraries from CR, we apply KNN retrieval to each image in the training set to generate pseudo-masks, which could be used to train the model (\cref{sec:mvkr}). 

\subsection{Clustering-then-Retrieval}\label{sec:cr}
Unlike retrieval-augmented segmentation~\cite{karazija24diffusion}, which uses diffusion models to generate prototypes, we aim to mine domain-adapted prototypes directly from the COD dataset.
However, in the absence of annotations, the key challenge lies in distinguishing between foreground and background to ensure that the resulting prototype library contains minimal noise.
To this end, we propose CR to mine high-quality prototypes from the dataset. CR first employs spectral clustering to generate coarse masks, which roughly distinguish camouflaged objects from the environment. It then uses mask-averaged pooling to obtain global features that characterize either the foreground or the background. Based on these global features, CR employs retrieval to select appropriate local features as the prototypes.

\noindent\textbf{Spectral clustering}
The first step for CR is to employ spectral clustering to generate coarse masks.
Compared to clustering high-dimensional features directly, spectral clustering maps the data to a low-dimensional space using feature decomposition by constructing the feature similarity matrix and Laplacian matrix, and then clustering using KMeans.

Given the image $\mathbf{I}\in\mathbb{R}^{H\times W \times 3}$ from COD training set, we utilize self-supervised model DINOv2~\cite{oquab2023dinov2} to extract the feature map $\mathbf{F}\in\mathbb{R}^{h \times w \times d}$, where $d$ denotes feature dimension. We then construct an undirected graph $\mathcal{G}=(\mathcal{V}, \mathcal{E})$ based on the feature maps, with local features $\mathbf{F_{i,j}\in\mathbb{R}^d}$ as nodes and cosine similarity between features as edges.
After getting the flattened feature map $\mathbf{F^\prime}\in\mathbb{R}^{hw\times d}$, the adjacency matrix $\mathbf{W}$ of the graph $\mathcal{G}$ is formulated as
\begin{equation}
    \mathbf{W_{i,j}} = \frac{\mathbf{F^\prime_i}\ \cdot \mathbf{F^\prime_j}}{\Vert \mathbf{F^\prime_i}\Vert \Vert \mathbf{F^\prime_j}\Vert}.
\end{equation}
In this paper, we consider negative similarity as 0,
\begin{equation}
    \mathbf{W_{i,j}} = \max(\mathbf{W_{i,j}}, 0).
\end{equation}

The Laplacian matrix for $\mathcal{G}$ is then given by $\mathbf{D}-\mathbf{W}$, where $\mathbf{D}$ with $\mathbf{D_{i,i}}=\sum_{j}\mathbf{W_{i,j}}$ indicates the diagonal matrix. The normalized Laplacian could be derived by
\begin{equation}
    \mathbf{L}=\mathbf{D^{-\frac{1}{2}}}(\mathbf{D}-\mathbf{W})\mathbf{D^{-\frac{1}{2}}}.
\end{equation}

We take the eigenvectors of $\mathbf{L}$ as new features and adopt KMeans to cluster the features into two clusters.
A simple prior is then adopted to assign foreground or background categories to these two clusters: the foreground is usually at the center of the image, and thus occupies a lower proportion of border pixels compared to the background. In this way, we obtain the coarse mask $\mathbf{M}\in\mathbb{R}^{h\times w}$.

\noindent\textbf{Cross-category retrieval}
The second step for CR is to employ a retrieval-based method to extract prototypes from images.
Given the mask $\mathbf{M}$ from spectral clustering, We employ mask-averaged pooling to obtain the global features for the foreground and background, respectively.
\begin{gather}
    \mathbf{F^g_f} = \frac{\sum_{i,j}\mathbf{F\odot \mathbf{M}}}{\sum_{i,j}\mathbf{M}},\\
    \mathbf{F^g_b} = \frac{\sum_{i,j}\mathbf{F\odot (1-\mathbf{M}})}{\sum_{i,j}(1-\mathbf{M})}.
\end{gather}
It is notable that despite some noise in M, average pooling enables each global feature to roughly characterize the corresponding category. We then derive the foreground and background feature sets by
\begin{gather}
    \mathbf{S_f}=\{\mathbf{F_{i,j}} \mid \mathbf{M_{i,j}=1}, 1\le i \le h,1\le j \le w\},\\
    \mathbf{S_b}=\{\mathbf{F_{i,j}} \mid \mathbf{M_{i,j}=0}, 1\le i \le h,1\le j \le w\}.
\end{gather}

To select the foreground prototype, we identify the feature from the foreground set $\mathbf{S_f}$ that is least similar to the global background feature $\mathbf{F^g_b}$. This can be expressed as:
\begin{equation}
    \mathbf{P^f} =  \arg \min_{\mathbf{\mathbf{s}} \in \mathbf{S_f}} \left( \frac{\mathbf{s} \cdot \mathbf{F^g_b}}{\Vert\mathbf{\mathbf{s}}\Vert \Vert\mathbf{F^g_b}\Vert} \right).
\end{equation}

Notably, we do not select the foreground prototype based on the most similar feature to the global foreground feature $\mathbf{F^g_f}$. This design choice is intentional: by choosing the foreground feature least similar to the global background feature, we enhance the distinction between the foreground and background prototypes. This differentiation is crucial for distinguishing camouflaged objects from the environment during the KNN retrieval stage. Our ablation experiments in~\cref{ablation} show that this cross-category retrieval strategy significantly improves performance.

Similarly, the background prototype is chosen by selecting the feature that is least similar to the global foreground feature from the background feature set:
\begin{equation}
    \mathbf{P^b} =  \arg \min_{\mathbf{\mathbf{s}} \in \mathbf{S_b}} \left( \frac{\mathbf{s} \cdot \mathbf{F^g_f}}{\Vert\mathbf{\mathbf{s}}\Vert \Vert\mathbf{F^g_f}\Vert} \right).
\end{equation}

\begin{figure}[!t]
\centering
\begin{center}
\includegraphics[width=\linewidth]{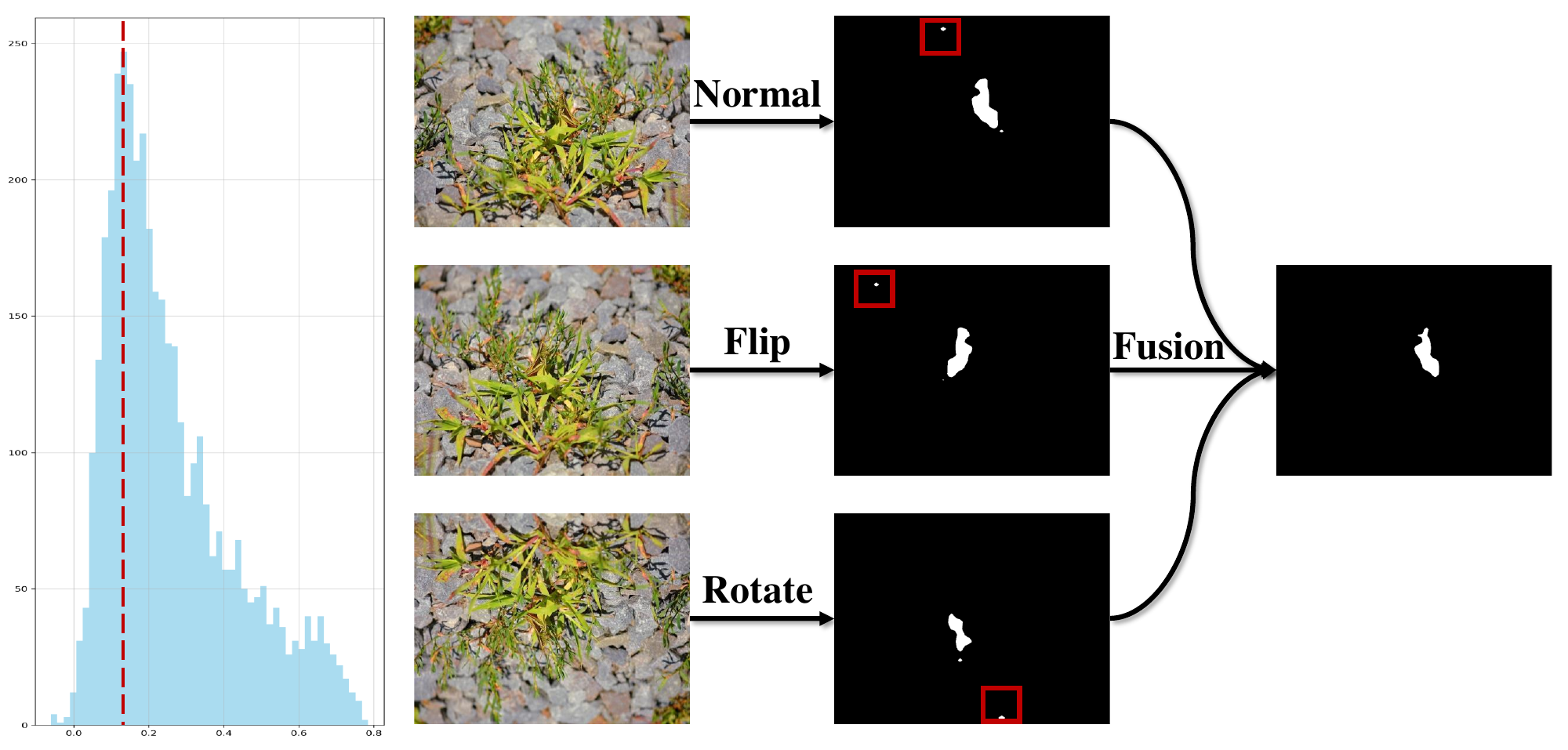}
\end{center}
\vspace{-1em}
\caption{\textbf{LEFT:} Histogram of the global feature similarity distribution across all images in the dataset. In this study, we focus only on the similarity values to the left of the peak (indicated by the red dashed line). \textbf{RIGHT:} Multi-View KNN Retrieval. To mitigate noisy retrieval results (highlighted in the red box) caused by artifacts in the feature map, we combine the masks generated by transforming the image through different viewpoints, resulting in a final fused mask.
}
\vspace{-1em}
\label{mvkr}
\end{figure}

\noindent\textbf{Histogram-based image filtering}
The prototype libraries are constructed by aggregating prototypes from each image in the dataset. However, since spectral clustering may yield inaccurate segments for certain images, the prototypes from these images may not accurately represent the corresponding categories. Given that erroneous clustering reduces the distinction between foreground and background, thereby increasing the similarity between their global features, we propose a histogram-based adaptive thresholding method to filter out images with high similarity.

We first employ spectral clustering to generate coarse masks for each image in the dataset and calculate the global feature similarity $\frac{\mathbf{F^g_f}\cdot\mathbf{F^g_b}}{\Vert \mathbf{F^g_f} \Vert \Vert \mathbf{F^g_b}\Vert}$ for each image based on these masks. The histogram of these similarities is shown on the left of~\cref{mvkr}. The distribution roughly forms a single peak. In this paper, we select the similarity corresponding to the peak's vertex as the adaptive threshold, considering only images with a similarity below this threshold. We denote the final foreground and background prototype libraries by $\{\mathbf{P^f_i}\}^n_{i=1}$ and $\{\mathbf{P^b_i}\}^n_{i=1}$, respectively, where $n$ indicates the number of prototypes.

\subsection{Multi-View KNN Retrieval}\label{sec:mvkr}
In this part, we introduce using KNN retrieval to generate pseudo-masks for each image in the dataset based on prototype libraries from CR.
Similarly, given the image $\mathbf{I}\in\mathbb{R}^{H\times W \times 3}$, we first extract its feature map $\mathbf{F}\in\mathbb{R}^{h \times w \times d}$ using DINOv2. For each local feature $\mathbf{F_{i,j}}\in\mathbb{R}^{d}$, we apply KNN to find the top-$K$ most similar prototypes from $\{\mathbf{P^f_i}\}^n_{i=1}$ and $\{\mathbf{P^b_i}\}^n_{i=1}$, using cosine similarity to measure the similarity between $\mathbf{F_{i,j}}$ and the prototypes. To classify each feature $\mathbf{F_{i,j}}$ as foreground or background, we use a voting mechanism over the top-$K$ retrieved prototypes. This process is repeated for all features in $\mathbf{F}$ , yielding the pseudo-mask $\mathbf{m}\in\mathbb{R}^{h\times w}$. After upsampling, the final mask matching the resolution of $\mathbf{I}$ could be obtained.

However, artifacts (features that do not reflect the true semantics of the image) in the feature maps may introduce noise into the retrieval results. Since the location of these artifacts varies with the image's viewpoint, we propose a Multi-View KNN Retrieval (MVKR) method that does not require fine-tuning. MVKR reduces the impact of artifacts by combining retrieval results from images derived from different viewpoints.

As illustrated on the right of~\cref{mvkr}, we first generate multiple views of the same image by applying transformations such as flipping or rotating. For each transformed image, we perform KNN retrieval to obtain the corresponding mask for the given viewpoint. To recover the mask for the normal view, we apply an inverse transformation. Finally, we use a voting mechanism to combine these masks into a single, unified result.

\section{Experiment}
\subsection{Experimental setup}
\noindent \textbf{Datasets and evaluation metrics}
Following previous works, our training dataset consists of 3,040 images from COD10K~\cite{ref1} and 1,000 images from CAMO~\cite{camo}. We test our method on four widely used benchmarks: CHAMELEON (76 test images)~\cite{chameleon}, CAMO (250 test images), COD10K (2,026 test images), and NC4K (4,121 test images)~\cite{lv2021simultaneously}. We employ four metrics for evaluation, including structure measure ($S_\alpha$)~\cite{structuremeasure}, mean E-measure ($E_\phi$)~\cite{emean}, weighted F-measure ($F^\omega_\beta$)~\cite{weightf} and mean absolute error ($M$)~\cite{mae}.

\noindent \textbf{Implementation details}
In this paper, we focus on generating pseudo-masks for training COD models, rather than designing the models themselves. We choose SINet-V2~\cite{ref2} as the training model, and the whole training process is identical to SINet-V2. We employ the self-supervised model DINOv2-ViT-L14~\cite{oquab2023dinov2} as the feature extractor and extract features from the last layer. During Multi-View KNN Retrieval, we retrieve the top-$K$ most similar prototypes from the prototype libraries, with $K$ set to 512. To generate multiple views of each image, we apply horizontal and vertical flipping, as well as rotations ($90^\circ$, $180^\circ$, and $270^\circ$). The FAISS library~\cite{faiss} is used for efficient retrieval. All images are resized to $476\times476$.

\begin{table*}[!t]
\resizebox{\textwidth}{!}{%
\begin{tabular}{lllcccccccccccccccc}
\hline
\multicolumn{1}{l|}{} & \multicolumn{1}{l|}{} & \multicolumn{1}{l|}{} & \multicolumn{4}{c|}{\textbf{CHAMELEON}} & \multicolumn{4}{c|}{\textbf{CAMO}} & \multicolumn{4}{c|}{\textbf{COD10K}} & \multicolumn{4}{c}{\textbf{NC4K}} \\ \cline{4-19} 
\multicolumn{1}{l|}{\multirow{-2}{*}{\textbf{Method}}} & \multicolumn{1}{l|}{\multirow{-2}{*}{\textbf{Feat. Extra. / SAM}}} & \multicolumn{1}{l|}{\multirow{-2}{*}{\textbf{Venue}}} & \multicolumn{1}{l}{\textbf{$S_\alpha\uparrow$}} & \multicolumn{1}{l}{\textbf{$E_\phi\uparrow$}} & \multicolumn{1}{l}{\textbf{$F^\omega_\beta\uparrow$}} & \multicolumn{1}{l|}{\textbf{$M\downarrow$}} & \multicolumn{1}{l}{\textbf{$S_\alpha\uparrow$}} & \multicolumn{1}{l}{\textbf{$E_\phi\uparrow$}} & \multicolumn{1}{l}{\textbf{$F^\omega_\beta\uparrow$}} & \multicolumn{1}{l|}{\textbf{$M\downarrow$}} & \multicolumn{1}{l}{\textbf{$S_\alpha\uparrow$}} & \multicolumn{1}{l}{\textbf{$E_\phi\uparrow$}} & \multicolumn{1}{l}{\textbf{$F^\omega_\beta\uparrow$}} & \multicolumn{1}{l|}{\textbf{$M\downarrow$}} & \multicolumn{1}{l}{\textbf{$S_\alpha\uparrow$}} & \multicolumn{1}{l}{\textbf{$E_\phi\uparrow$}} & \multicolumn{1}{l}{\textbf{$F^\omega_\beta\uparrow$}} & \multicolumn{1}{l}{\textbf{$M\downarrow$}} \\ \hline
\multicolumn{19}{l}{\textbf{Unsupervised Segmentaion}} \\ \hline
\multicolumn{1}{l|}{LOST} & \multicolumn{1}{l|}{DINO-ViT-S16} & \multicolumn{1}{l|}{BMVC21} & 0.631 & 0.713 & 0.418 & \multicolumn{1}{c|}{0.145} & 0.644 & 0.688 & 0.450 & \multicolumn{1}{c|}{0.166} & 0.673 & 0.722 & 0.433 & \multicolumn{1}{c|}{0.095} & 0.705 & 0.748 & 0.532 & 0.118 \\
\multicolumn{1}{l|}{DeepSpectral} & \multicolumn{1}{l|}{DINO-ViT-S8} & \multicolumn{1}{l|}{CVPR22} & 0.662 & 0.677 & 0.465 & \multicolumn{1}{c|}{0.187} & 0.640 & 0.659 & 0.457 & \multicolumn{1}{c|}{0.215} & 0.621 & 0.625 & 0.363 & \multicolumn{1}{c|}{0.162} & 0.719 & 0.739 & 0.542 & 0.133 \\
\multicolumn{1}{l|}{DeepSpectral} & \multicolumn{1}{l|}{DINOv2-ViT-L14} & \multicolumn{1}{l|}{CVPR22} & 0.605 & 0.613 & 0.402 & \multicolumn{1}{c|}{0.225} & 0.560 & 0.563 & 0.366 & \multicolumn{1}{c|}{0.292} & 0.493 & 0.472 & 0.227 & \multicolumn{1}{c|}{0.314} & 0.588 & 0.588 & 0.378 & 0.255 \\
\multicolumn{1}{l|}{TokenCut} & \multicolumn{1}{l|}{DINO-ViT-S16} & \multicolumn{1}{l|}{CVPR22} & 0.687 & 0.757 & 0.516 & \multicolumn{1}{c|}{0.115} & 0.661 & 0.717 & 0.498 & \multicolumn{1}{c|}{0.157} & 0.682 & 0.729 & 0.468 & \multicolumn{1}{c|}{0.095} & 0.752 & 0.804 & 0.614 & 0.093 \\
\multicolumn{1}{l|}{TokenCut} & \multicolumn{1}{l|}{DINOv2-ViT-L14} & \multicolumn{1}{l|}{CVPR22} & 0.708 & 0.747 & 0.510 & \multicolumn{1}{c|}{0.078} & 0.608 & 0.612 & 0.385 & \multicolumn{1}{c|}{0.148} & 0.637 & 0.672 & 0.370 & \multicolumn{1}{c|}{0.077} & 0.697 & 0.731 & 0.511 & 0.095 \\
\multicolumn{1}{l|}{MaskCut} & \multicolumn{1}{l|}{DINO-ViT-B8} & \multicolumn{1}{l|}{CVPR23} & 0.636 & 0.662 & 0.448 & \multicolumn{1}{c|}{0.194} & 0.646 & 0.653 & 0.454 & \multicolumn{1}{c|}{0.206} & 0.626 & 0.628 & 0.363 & \multicolumn{1}{c|}{0.170} & 0.708 & 0.717 & 0.518 & 0.152 \\
\multicolumn{1}{l|}{MaskCut} & \multicolumn{1}{l|}{DINOv2-ViT-L14} & \multicolumn{1}{l|}{CVPR23} & 0.653 & 0.665 & 0.462 & \multicolumn{1}{c|}{0.169} & 0.628 & 0.656 & 0.439 & \multicolumn{1}{c|}{0.200} & 0.607 & 0.624 & 0.346 & \multicolumn{1}{c|}{0.165} & 0.683 & 0.710 & 0.498 & 0.152 \\
\multicolumn{1}{l|}{FOUND} & \multicolumn{1}{l|}{DINO-ViT-S8} & \multicolumn{1}{l|}{CVPR23} & 0.609 & 0.616 & 0.374 & \multicolumn{1}{c|}{0.215} & 0.572 & 0.571 & 0.379 & \multicolumn{1}{c|}{0.280} & 0.522 & 0.503 & 0.252 & \multicolumn{1}{c|}{0.247} & 0.610 & 0.613 & 0.399 & 0.216 \\
\multicolumn{1}{l|}{FOUND} & \multicolumn{1}{l|}{DINOv2-ViT-L14} & \multicolumn{1}{l|}{CVPR23} & 0.656 & 0.731 & 0.459 & \multicolumn{1}{c|}{0.115} & 0.590 & 0.635 & 0.378 & \multicolumn{1}{c|}{0.166} & 0.636 & 0.710 & 0.394 & \multicolumn{1}{c|}{0.088} & 0.669 & 0.734 & 0.483 & 0.113 \\
\multicolumn{1}{l|}{ProMerge} & \multicolumn{1}{l|}{DINO-ViT-B8} & \multicolumn{1}{l|}{ECCV24} & 0.712 & 0.757 & 0.534 & \multicolumn{1}{c|}{0.087} & 0.692 & 0.732 & 0.542 & \multicolumn{1}{c|}{0.134} & 0.714 & 0.756 & 0.504 & \multicolumn{1}{c|}{0.078} & 0.773 & 0.817 & 0.639 & 0.081 \\
\multicolumn{1}{l|}{ProMerge} & \multicolumn{1}{l|}{DINOv2-ViT-L14} & \multicolumn{1}{l|}{ECCV24} & 0.741 & 0.787 & 0.567 & \multicolumn{1}{c|}{0.085} & 0.679 & 0.706 & 0.502 & \multicolumn{1}{c|}{0.142} & 0.674 & 0.714 & 0.435 & \multicolumn{1}{c|}{0.081} & 0.726 & 0.771 & 0.558 & 0.096 \\
\multicolumn{1}{l|}{VoteCut} & \multicolumn{1}{l|}{DINO(v2) Ensemble} & \multicolumn{1}{l|}{CVPR24} & 0.699 & 0.763 & 0.556 & \multicolumn{1}{c|}{0.138} & 0.663 & 0.708 & 0.506 & \multicolumn{1}{c|}{0.158} & 0.702 & 0.763 & 0.504 & \multicolumn{1}{c|}{0.092} & 0.756 & 0.812 & 0.628 & 0.094 \\
\multicolumn{1}{l|}{VoteCut} & \multicolumn{1}{l|}{DINOv2-ViT-L14} & \multicolumn{1}{l|}{CVPR24} & 0.679 & 0.695 & 0.484 & \multicolumn{1}{c|}{0.110} & 0.580 & 0.572 & 0.356 & \multicolumn{1}{c|}{0.164} & 0.645 & 0.673 & 0.390 & \multicolumn{1}{c|}{0.082} & 0.674 & 0.694 & 0.476 & 0.104 \\
\multicolumn{1}{l|}{DiffCut} & \multicolumn{1}{l|}{SSD-1B} & \multicolumn{1}{l|}{NeurIPS24} & 0.574 & 0.613 & 0.390 & \multicolumn{1}{c|}{0.220} & 0.627 & 0.662 & 0.454 & \multicolumn{1}{c|}{0.185} & 0.628 & 0.667 & 0.372 & \multicolumn{1}{c|}{0.120} & 0.693 & 0.739 & 0.514 & 0.122 \\
\rowcolor[HTML]{EFEFEF} 
\multicolumn{1}{l|}{\cellcolor[HTML]{EFEFEF}RISE} & \multicolumn{1}{l|}{\cellcolor[HTML]{EFEFEF}DINO-ViT-S16} & \multicolumn{1}{l|}{\cellcolor[HTML]{EFEFEF}-} & 0.670 & 0.738 & 0.489 & \multicolumn{1}{c|}{\cellcolor[HTML]{EFEFEF}0.095} & 0.655 & 0.712 & 0.494 & \multicolumn{1}{c|}{\cellcolor[HTML]{EFEFEF}0.135} & 0.716 & 0.790 & 0.518 & \multicolumn{1}{c|}{\cellcolor[HTML]{EFEFEF}0.063} & 0.754 & 0.814 & 0.627 & 0.083 \\
\rowcolor[HTML]{EFEFEF} 
\multicolumn{1}{l|}{\cellcolor[HTML]{EFEFEF}RISE} & \multicolumn{1}{l|}{\cellcolor[HTML]{EFEFEF}DINO-ViT-B16} & \multicolumn{1}{l|}{\cellcolor[HTML]{EFEFEF}-} & 0.679 & 0.741 & 0.508 & \multicolumn{1}{c|}{\cellcolor[HTML]{EFEFEF}0.090} & 0.664 & 0.715 & 0.507 & \multicolumn{1}{c|}{\cellcolor[HTML]{EFEFEF}0.132} & 0.724 & 0.795 & 0.526 & \multicolumn{1}{c|}{\cellcolor[HTML]{EFEFEF}0.062} & 0.759 & 0.818 & 0.633 & 0.081 \\
\rowcolor[HTML]{EFEFEF} 
\multicolumn{1}{l|}{\cellcolor[HTML]{EFEFEF}RISE} & \multicolumn{1}{l|}{\cellcolor[HTML]{EFEFEF}DINOv2-ViT-S14} & \multicolumn{1}{l|}{\cellcolor[HTML]{EFEFEF}-} & 0.762 & 0.810 & 0.631 & \multicolumn{1}{c|}{\cellcolor[HTML]{EFEFEF}0.068} & 0.684 & 0.726 & 0.532 & \multicolumn{1}{c|}{\cellcolor[HTML]{EFEFEF}0.127} & 0.741 & 0.814 & 0.564 & \multicolumn{1}{c|}{\cellcolor[HTML]{EFEFEF}0.059} & 0.788 & 0.847 & 0.674 & 0.069 \\
\rowcolor[HTML]{EFEFEF} 
\multicolumn{1}{l|}{\cellcolor[HTML]{EFEFEF}RISE} & \multicolumn{1}{l|}{\cellcolor[HTML]{EFEFEF}DINOv2-ViT-B14} & \multicolumn{1}{l|}{\cellcolor[HTML]{EFEFEF}-} & {\ul 0.805} & {\ul 0.858} & {\ul 0.675} & \multicolumn{1}{c|}{\cellcolor[HTML]{EFEFEF}{\ul 0.052}} & {\ul 0.722} & {\ul 0.775} & {\ul 0.587} & \multicolumn{1}{c|}{\cellcolor[HTML]{EFEFEF}{\ul 0.113}} & {\ul 0.753} & {\ul 0.827} & {\ul 0.578} & \multicolumn{1}{c|}{\cellcolor[HTML]{EFEFEF}{\ul 0.053}} & {\ul 0.797} & {\ul 0.860} & {\ul 0.687} & {\ul 0.064} \\
\rowcolor[HTML]{EFEFEF} 
\multicolumn{1}{l|}{\cellcolor[HTML]{EFEFEF}RISE} & \multicolumn{1}{l|}{\cellcolor[HTML]{EFEFEF}DINOv2-ViT-L14} & \multicolumn{1}{l|}{\cellcolor[HTML]{EFEFEF}-} & \textbf{0.822} & \textbf{0.884} & \textbf{0.720} & \multicolumn{1}{c|}{\cellcolor[HTML]{EFEFEF}\textbf{0.050}} & \textbf{0.734} & \textbf{0.787} & \textbf{0.610} & \multicolumn{1}{c|}{\cellcolor[HTML]{EFEFEF}\textbf{0.109}} & \textbf{0.763} & \textbf{0.840} & \textbf{0.600} & \multicolumn{1}{c|}{\cellcolor[HTML]{EFEFEF}\textbf{0.049}} & \textbf{0.805} & \textbf{0.868} & \textbf{0.705} & \textbf{0.061} \\ \hline
\multicolumn{19}{l}{\textbf{Prompt-based Segmentation}} \\ \hline
\multicolumn{1}{l|}{WS-SAM*} & \multicolumn{1}{l|}{SAM-ViT-H} & \multicolumn{1}{l|}{NeurIPS23} & {\ul 0.795} & 0.824 & {\ul 0.676} & \multicolumn{1}{c|}{0.099} & \textbf{0.781} & {\ul 0.807} & \textbf{0.658} & \multicolumn{1}{c|}{0.108} & {\ul 0.787} & {\ul 0.838} & {\ul 0.622} & \multicolumn{1}{c|}{0.057} & \textbf{0.829} & {\ul 0.867} & {\ul 0.727} & {\ul 0.063} \\
\multicolumn{1}{l|}{GenSAM} & \multicolumn{1}{l|}{SAM-ViT-H} & \multicolumn{1}{l|}{AAAI24} & 0.659 & 0.716 & 0.495 & \multicolumn{1}{c|}{0.153} & 0.633 & 0.673 & 0.458 & \multicolumn{1}{c|}{0.188} & 0.641 & 0.675 & 0.390 & \multicolumn{1}{c|}{0.136} & 0.702 & 0.744 & 0.524 & 0.128 \\
\multicolumn{1}{l|}{ProMac} & \multicolumn{1}{l|}{SAM-ViT-H} & \multicolumn{1}{l|}{NeurIPS24} & 0.786 & {\ul 0.842} & 0.665 & \multicolumn{1}{c|}{{\ul 0.066}} & 0.754 & \textbf{0.812} & 0.645 & \multicolumn{1}{c|}{\textbf{0.101}} & 0.774 & 0.835 & 0.609 & \multicolumn{1}{c|}{{\ul 0.052}} & 0.812 & 0.862 & 0.711 & 0.065 \\
\rowcolor[HTML]{EFEFEF} 
\multicolumn{1}{l|}{\cellcolor[HTML]{EFEFEF}RISE} & \multicolumn{1}{l|}{\cellcolor[HTML]{EFEFEF}SAM-ViT-H} & \multicolumn{1}{l|}{\cellcolor[HTML]{EFEFEF}-} & \textbf{0.823} & \textbf{0.882} & \textbf{0.733} & \multicolumn{1}{c|}{\cellcolor[HTML]{EFEFEF}\textbf{0.055}} & {\ul 0.760} & {\ul 0.807} & {\ul 0.651} & \multicolumn{1}{c|}{\cellcolor[HTML]{EFEFEF}{\ul 0.102}} & \textbf{0.790} & \textbf{0.854} & \textbf{0.643} & \multicolumn{1}{c|}{\cellcolor[HTML]{EFEFEF}\textbf{0.044}} & {\ul 0.825} & \textbf{0.874} & \textbf{0.736} & \textbf{0.056} \\ \hline
\end{tabular}%
}
\vspace{-1em}
\caption{Comparisons with unsupervised and prompt-based segmentation methods across four commonly used benchmark datasets. ``$\uparrow$/$\downarrow$": The higher/lower the better. ``*": The prompts for SAM are derived by manual labeling. The best and second-best results are \textbf{bolded} and \underline{underlined} to highlight, respectively.}
\vspace{-1em}
\label{quantitative_comparison}
\end{table*}

\subsection{Comparisons with state-of-the-arts}
\noindent \textbf{Comparison methods}
We compare our method with both unsupervised and prompt-based segmentation approaches. For the unsupervised setting, we re-implement eight state-of-the-art methods on COD using their official codes. These include seven DINO(v2)-based methods (DeepSpectral~\cite{deepspectral}, LOST~\cite{LOST}, FOUND~\cite{FOUND}, TokenCut~\cite{tokencut}, MaskCut~\cite{cutler}, VoteCut~\cite{CuVLER}, ProMerge~\cite{promerge}), and one diffusion model-based method, DiffCut~\cite{diffcut}. For the DINO(v2)-based methods, following CuVLER~\cite{CuVLER}, we resize images to $480\times480$ for DINO and $476\times476$ for DINOv2. To ensure a fair comparison, we remove all post-processing techniques, such as bilateral solvers~\cite{bilateralsolver} and conditional random fields~\cite{crf}. For methods like MaskCut that produce multiple masks, we combine them into a single binary mask. We also observe that some DINO-based methods, such as ProMerge, perform poorly with DINOv2 as the feature extractor. In this situation, we tune the main hyperparameters and experiment with features at different locations (Query, Key, Value, QKV, and last layer) to optimize performance.
For prompt-based setting, we integrate our method with SAM~\cite{sam} by extracting bounding boxes from pseudo-masks, which serve as prompts for SAM. We select three methods for comparisons: GenSAM~\cite{gensam}, ProMac~\cite{ProMaC}, and WS-SAM~\cite{wssam}. For GenSAM and ProMac, we use their official codes to generate pseudo-labels on the training set. 
It is notable that for ProMac we adopt LLaVA-1.5-7B~\cite{llava1.5} as LMM instead of LLaVA-1.5-13B due to limited GPU memory.
For WS-SAM, we use the pseudo-masks provided by the authors.

\noindent\textbf{Quantitative comparison}
In comparison to unsupervised methods, as shown in \cref{quantitative_comparison}, RISE outperforms all state-of-the-art techniques. On the COD10K dataset, our method achieves a minimum improvement of 8\% and 9\% in metrics $E_\phi$ and $F^\omega_\beta$, respectively. Notably, RISE delivers top-tier performance on the more challenging COD10K and NC4K datasets, regardless of the DINO configuration. These results highlight the advantage of our approach, which leverages dataset-level information to retrieve camouflaged objects, in contrast to traditional unsupervised methods that focus on intra-image similarity for foreground segmentation.
When compared to prompt-based segmentation methods, our approach also demonstrates superior performance. Despite WS-SAM~\cite{wssam} providing prompts for SAM based on weakly supervised signals from manually labeled data, RISE outperforms it in most metrics, showcasing its enhanced ability to localize camouflaged objects. Additionally, GenSAM~\cite{gensam} and ProMac~\cite{ProMaC} leverage multimodal LLMs or diffusion models to generate prompts. However, due to the high parameter intensity of these models, generating pseudo-masks for the entire training dataset can take days. In contrast, RISE uses only self-supervised models, significantly reducing inference time.

\begin{figure*}[!t]
\centering
\begin{center}
\includegraphics[width=\linewidth]{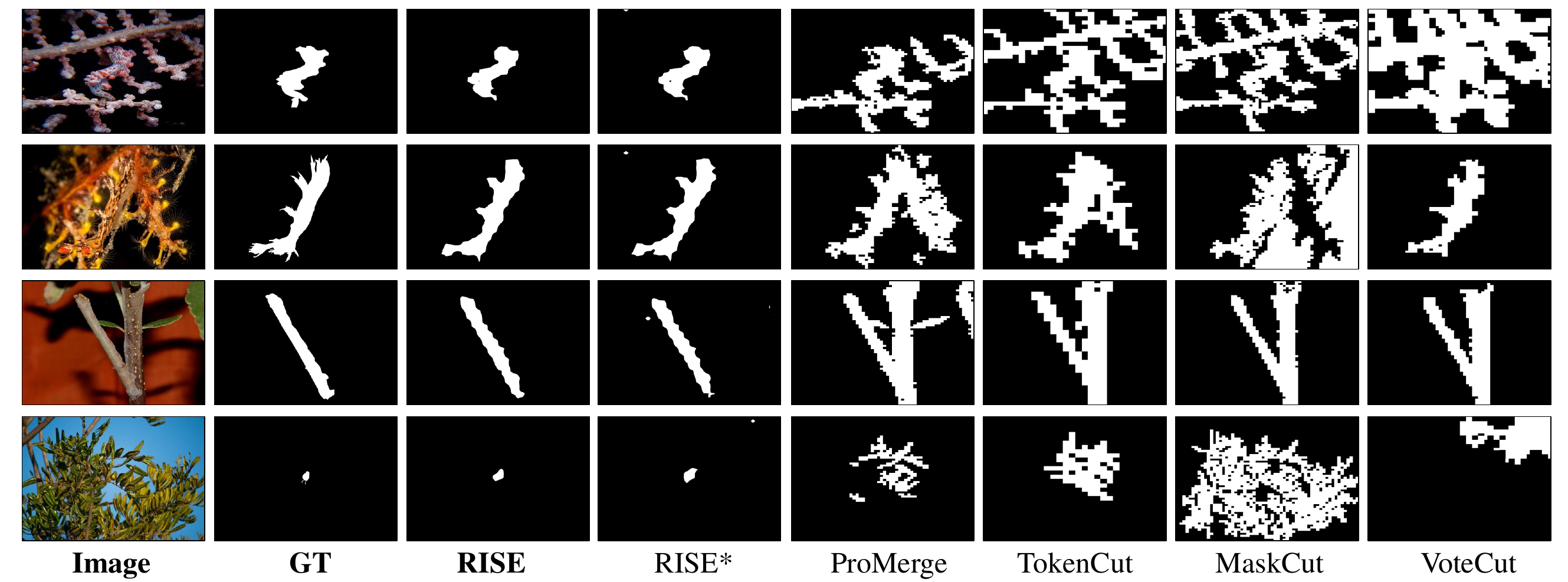}
\end{center}
\vspace{-1em}
\caption{Qualitative comparisons with four SOTA unsupervised methods. RISE* denotes RISE without Multi-View Retrieval.}
\label{qualitative_comparison}
\vspace{-1em}
\end{figure*}

\begin{table*}[!t]
\resizebox{\textwidth}{!}{%
\begin{tabular}{@{}c|l|l|cccc|cccc|cccc|cccc@{}}
\toprule
 &  &  & \multicolumn{4}{c|}{\textbf{CHAMELEON}} & \multicolumn{4}{c|}{\textbf{CAMO}} & \multicolumn{4}{c|}{\textbf{COD10K}} & \multicolumn{4}{c}{\textbf{NC4K}} \\ \cmidrule(l){4-19} 
\multirow{-2}{*}{\textbf{Row index}} & \multirow{-2}{*}{\textbf{Component}} & \multirow{-2}{*}{\textbf{Ablation}} & \multicolumn{1}{l}{\textbf{$S_\alpha\uparrow$}} & \multicolumn{1}{l}{\textbf{$E_\phi\uparrow$}} & \multicolumn{1}{l}{\textbf{$F^\omega_\beta\uparrow$}} & \multicolumn{1}{l|}{\textbf{$M\downarrow$}} & \multicolumn{1}{l}{\textbf{$S_\alpha\uparrow$}} & \multicolumn{1}{l}{\textbf{$E_\phi\uparrow$}} & \multicolumn{1}{l}{\textbf{$F^\omega_\beta\uparrow$}} & \multicolumn{1}{l|}{\textbf{$M\downarrow$}} & \multicolumn{1}{l}{\textbf{$S_\alpha\uparrow$}} & \multicolumn{1}{l}{\textbf{$E_\phi\uparrow$}} & \multicolumn{1}{l}{\textbf{$F^\omega_\beta\uparrow$}} & \multicolumn{1}{l|}{\textbf{$M\downarrow$}} & \multicolumn{1}{l}{\textbf{$S_\alpha\uparrow$}} & \multicolumn{1}{l}{\textbf{$E_\phi\uparrow$}} & \multicolumn{1}{l}{\textbf{$F^\omega_\beta\uparrow$}} & \multicolumn{1}{l}{\textbf{$M\downarrow$}} \\ \midrule
a & Paradigm & w image-level modeling & 0.743 & 0.782 & 0.607 & 0.133 & 0.665 & 0.690 & 0.506 & 0.197 & 0.641 & 0.662 & 0.414 & 0.169 & 0.727 & 0.760 & 0.569 & 0.134 \\ \midrule
b &  & w/o Cross-Category Retrieval & 0.787 & 0.851 & 0.660 & 0.063 & 0.743 & 0.808 & 0.615 & 0.107 & 0.710 & 0.781 & 0.518 & 0.065 & 0.764 & 0.833 & 0.639 & 0.076 \\
c & \multirow{-2}{*}{CR} & w/o Histogram-based Filtering & 0.820 & 0.878 & 0.718 & 0.052 & 0.733 & 0.791 & 0.609 & 0.112 & 0.744 & 0.822 & 0.575 & 0.055 & 0.793 & 0.859 & 0.688 & 0.065 \\ \midrule
d & KNN Retrieval & w/o Multi-View Retrieval & 0.829 & 0.885 & 0.719 & 0.052 & 0.735 & 0.789 & 0.604 & 0.115 & 0.759 & 0.832 & 0.584 & 0.052 & 0.803 & 0.863 & 0.694 & 0.064 \\ \midrule
\rowcolor[HTML]{EFEFEF} 
e & Ours & - & 0.822 & 0.884 & 0.720 & 0.050 & 0.734 & 0.787 & 0.610 & 0.109 & 0.763 & 0.840 & 0.600 & 0.049 & 0.805 & 0.868 & 0.705 & 0.061 \\ \bottomrule
\end{tabular}%
}
\vspace{-1em}
\caption{Ablation experiments on the effect of each component.}
\label{ablation}
\vspace{-1em}
\end{table*}

\noindent\textbf{Qualitative comparison}
To illustrate RISE's ability to accurately localize and segment camouflaged objects in complex environments, we present a qualitative comparison with four leading unsupervised methods for generating pseudo-masks, as shown in~\cref{qualitative_comparison}. These methods, which rely on intra-image similarity, often fail to differentiate between targets and backgrounds with similar appearances, leading to suboptimal segmentation. In contrast, RISE overcomes this challenge by incorporating both foreground and background semantics at the dataset level, resulting in more effective separation. Notably, our method performs well in localizing small objects, as shown in the last row of~\cref{qualitative_comparison}.

\subsection{Ablation Study}
\noindent \textbf{Paradigm}
Our retrieval self-augmented pipeline combines a prototype generator (CR) with a KNN retrieval scheme. We first assess the effectiveness of our proposed paradigm for retrieving camouflaged objects using the COD dataset, comparing it to previous approaches that rely on single-image modeling. Spectral clustering, used in the CR stage, serves as the baseline. As shown in rows \textbf{a} and \textbf{e} of~\cref{ablation}, our method outperforms the baseline by more than 10\% on average across all datasets and metrics. This underscores the importance of fully utilizing dataset-level semantics to distinguish between highly similar foregrounds and backgrounds in a single image.

\noindent \textbf{Clustering-then-Retrieval}
The CR module incorporates two key components: cross-category retrieval and histogram-based image filtering, both designed to generate high-quality, distinguishable prototype libraries. As shown in rows \textbf{b} and \textbf{e}, incorporating cross-category retrieval improves RISE performance on the challenging COD10K dataset by 5.3\%, 5.9\%, and 8.2\% for the $S_\alpha$, $E_\phi$, and $F^\omega_\beta$ metrics, respectively. Furthermore, histogram-based image filtering not only boosts performance across all datasets (as shown in rows \textbf{c} and \textbf{e}) but also reduces the size of the prototype libraries, enhancing inference efficiency.

\noindent \textbf{Multi-View KNN Retrieval}
In the KNN retrieval stage, we address artifacts in the self-supervised model by aggregating retrieval results from multiple viewpoints. This not only improves performance (as indicated by rows \textbf{d} and \textbf{e}) but also ensures that noise-resistant results can be obtained without fine-tuning models, as demonstrated in the third and fourth columns of~\cref{qualitative_comparison} (best zoomed-in).

\begin{figure*}[!t]
\centering
\begin{center}
\includegraphics[width=\linewidth]{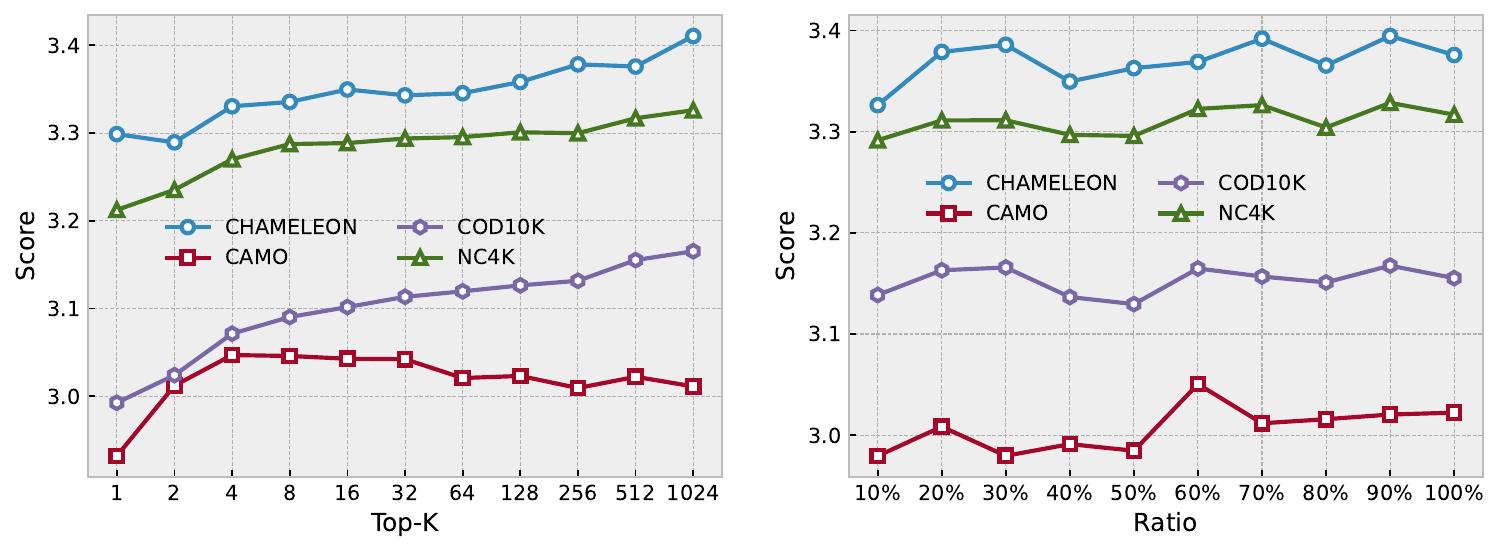}
\end{center}
\vspace{-2em}
\caption{\textbf{Left:} Sensitivity analysis of the top-$K$ hyperparameter; \textbf{Right:} Impact of the dataset size.}
\label{topk-and-datasetsize}
\vspace{-1em}
\end{figure*}

\begin{table}[!t]
\resizebox{\linewidth}{!}{%
\begin{tabular}{@{}l|cccc|cccc@{}}
\toprule
 & \multicolumn{4}{c|}{\textbf{COD10K}} & \multicolumn{4}{c}{\textbf{NC4K}} \\ \cmidrule(l){2-9} 
\multirow{-2}{*}{\textbf{Method}} & \multicolumn{1}{l}{\textbf{$S_\alpha\uparrow$}} & \multicolumn{1}{l}{\textbf{$E_\phi\uparrow$}} & \multicolumn{1}{l}{\textbf{$F^\omega_\beta\uparrow$}} & \multicolumn{1}{l|}{\textbf{$M\downarrow$}} & \multicolumn{1}{l}{\textbf{$S_\alpha\uparrow$}} & \multicolumn{1}{l}{\textbf{$E_\phi\uparrow$}} & \multicolumn{1}{l}{\textbf{$F^\omega_\beta\uparrow$}} & \multicolumn{1}{l}{\textbf{$M\downarrow$}} \\ \midrule
Previous SOTA & 0.714 & 0.756 & 0.504 & 0.078 & 0.773 & 0.817 & 0.639 & 0.081 \\ \midrule
KMeans & 0.400 & 0.404 & 0.155 & 0.410 & 0.494 & 0.514 & 0.303 & 0.343 \\
\rowcolor[HTML]{EFEFEF} 
KMeans+ & \textbf{0.723} & \textbf{0.788} & \textbf{0.536} & \textbf{0.073} & \textbf{0.777} & \textbf{0.836} & \textbf{0.656} & \textbf{0.078} \\ \midrule
GMM & 0.450 & 0.440 & 0.192 & 0.371 & 0.572 & 0.578 & 0.371 & 0.280 \\
\rowcolor[HTML]{EFEFEF} 
GMM+ & \textbf{0.719} & \textbf{0.789} & \textbf{0.534} & \textbf{0.069} & \textbf{0.771} & \textbf{0.835} & \textbf{0.652} & \textbf{0.078} \\ \midrule
HCA & 0.474 & 0.484 & 0.206 & 0.317 & 0.579 & 0.600 & 0.376 & 0.257 \\
\rowcolor[HTML]{EFEFEF} 
HCA+ & \textbf{0.733} & \textbf{0.805} & \textbf{0.556} & \textbf{0.064} & \textbf{0.786} & \textbf{0.848} & \textbf{0.672} & \textbf{0.071} \\ \midrule
Spectral & 0.641 & 0.662 & 0.414 & 0.169 & 0.727 & 0.760 & 0.569 & 0.134 \\
\rowcolor[HTML]{EFEFEF} 
Spectral+ & \textbf{0.763} & \textbf{0.840} & \textbf{0.600} & \textbf{0.049} & \textbf{0.805} & \textbf{0.868} & \textbf{0.705} & \textbf{0.061} \\ \bottomrule
\end{tabular}%
}
\vspace{-1em}
\caption{Experiments on different clustering methods. ``+": Integrating clustering methods with RISE.}
\label{abla-clustering}
\vspace{-1em}
\end{table}

\subsection{Further Analysis}
\noindent\textbf{Hyperparameter sensitivity}
Our method has only one hyperparameter: top-$K$. We adopt $\mathrm{Score}=S_\alpha+E_\phi+F^\omega_\beta+1-M$~\cite{senet} to indicate the overall performance. As shown in the left of~\cref{topk-and-datasetsize}, RISE performs well even with a small $K$. This robustness stems from the unique CR scheme, which reduces noise in the prototype libraries and enhances the distinction between different libraries.

\noindent\textbf{Dataset size}
In the right of~\cref{topk-and-datasetsize}, we investigate the impact of the dataset size. Leveraging CR's distinctive prototype mining mechanism, RISE demonstrates consistent and robust performance even when only 10\% of the training images (randomly selected) are used to construct the prototype libraries. This highlights the efficiency and effectiveness of RISE, particularly in scenarios with limited data.

\noindent\textbf{Clustering methods}
We explore several clustering methods, including KMeans, Gaussian Mixture Model (GMM), Hierarchical Clustering Analysis (HCA), and Spectral Clustering (Spectral). As shown in~\cref{abla-clustering}, while the performance of each method varies, integrating our approach significantly boosts the performance of all methods. 

\begin{table}[!t]
\resizebox{\linewidth}{!}{%
\begin{tabular}{@{}l|cccc|cccc@{}}
\toprule
 & \multicolumn{4}{c|}{\textbf{COD10K}} & \multicolumn{4}{c}{\textbf{NC4K}} \\ \cmidrule(l){2-9} 
\multirow{-2}{*}{\textbf{Method}} & \multicolumn{1}{l}{\textbf{$S_\alpha\uparrow$}} & \multicolumn{1}{l}{\textbf{$E_\phi\uparrow$}} & \multicolumn{1}{l}{\textbf{$F^\omega_\beta\uparrow$}} & \multicolumn{1}{l|}{\textbf{$M\downarrow$}} & \multicolumn{1}{l}{\textbf{$S_\alpha\uparrow$}} & \multicolumn{1}{l}{\textbf{$E_\phi\uparrow$}} & \multicolumn{1}{l}{\textbf{$F^\omega_\beta\uparrow$}} & \multicolumn{1}{l}{\textbf{$M\downarrow$}} \\ \midrule
LOST & 0.673 & 0.722 & 0.433 & 0.095 & 0.705 & 0.748 & 0.532 & 0.118 \\
\rowcolor[HTML]{EFEFEF} 
LOST+ & \textbf{0.766} & \textbf{0.854} & \textbf{0.615} & \textbf{0.046} & \textbf{0.799} & \textbf{0.869} & \textbf{0.708} & \textbf{0.061} \\ \midrule
DeepSpectral & 0.621 & 0.625 & 0.363 & 0.162 & 0.719 & 0.739 & 0.542 & 0.133 \\
\rowcolor[HTML]{EFEFEF} 
DeepSpectral+ & \textbf{0.765} & \textbf{0.842} & \textbf{0.605} & \textbf{0.050} & \textbf{0.804} & \textbf{0.860} & \textbf{0.702} & \textbf{0.063} \\ \midrule
TokenCut & 0.682 & 0.729 & 0.468 & 0.095 & 0.752 & 0.804 & 0.614 & 0.093 \\
\rowcolor[HTML]{EFEFEF} 
TokenCut+ & \textbf{0.769} & \textbf{0.848} & \textbf{0.607} & \textbf{0.048} & \textbf{0.813} & \textbf{0.876} & \textbf{0.715} & \textbf{0.059} \\ \midrule
MaskCut & 0.626 & 0.628 & 0.363 & 0.170 & 0.708 & 0.717 & 0.518 & 0.152 \\
\rowcolor[HTML]{EFEFEF} 
MaskCut+ & \textbf{0.763} & \textbf{0.848} & \textbf{0.608} & \textbf{0.047} & \textbf{0.793} & \textbf{0.856} & \textbf{0.698} & \textbf{0.063} \\ \midrule
ProMerge & 0.714 & 0.756 & 0.504 & 0.078 & 0.773 & 0.817 & 0.639 & 0.081 \\
\rowcolor[HTML]{EFEFEF} 
ProMerge+ & \textbf{0.770} & \textbf{0.850} & \textbf{0.622} & \textbf{0.045} & \textbf{0.805} & \textbf{0.868} & \textbf{0.715} & \textbf{0.059} \\ \midrule
VoteCut & 0.702 & 0.763 & 0.504 & 0.092 & 0.756 & 0.812 & 0.628 & 0.094 \\
\rowcolor[HTML]{EFEFEF} 
VoteCut+ & \textbf{0.766} & \textbf{0.845} & \textbf{0.608} & \textbf{0.048} & \textbf{0.807} & \textbf{0.869} & \textbf{0.710} & \textbf{0.061} \\ \bottomrule
\end{tabular}%
}
\vspace{-1em}
\caption{Experiments on the plug-and-play attribute of RISE. ``+": Integrating unsupervised methods with RISE.}
\label{tab:plugandplay}
\vspace{-1em}
\end{table}

\noindent\textbf{Plug-and-play}
By replacing spectral clustering in CR with unsupervised methods, our approach can be integrated with unsupervised methods.
This enhances the performance of these methods, as shown in~\cref{tab:plugandplay}.

\section{Conclusion}
In this paper, we offer a new perspective on camouflaged object detection: while a camouflaged object may be indistinguishable from its surroundings in a single image, it may become distinguishable when considered within the context of a dataset. Building on this insight, we introduce RISE, a retrieval self-augmented unsupervised COD paradigm. RISE is designed to separate hard-to-recognize targets within a single image by efficiently integrating and leveraging dataset-level information. To access this dataset-level information, we propose a Clustering-then-Retrieval approach. For segmenting camouflaged objects in a single image, we introduce Multi-View KNN Retrieval. Extensive experiments demonstrate that our method outperforms both unsupervised and prompt-based segmentation approaches.

\noindent\textbf{Acknowledgments.} This work was supported in part by The Hong Kong Polytechnic University under Grants P0048387, P0044520, P0050657, and P0049586, and in part by the Tianjin Science and Technology Major Project under Grant 24ZXZSSS00420.

{
    \small
    \bibliographystyle{ieeenat_fullname}
    \bibliography{main}
}

\end{document}